\documentclass[runningheads]{llncs}
\usepackage{graphicx}
\begin{document}
\title{Transformer-based end-to-end classification of variable-length volumetric data}
\titlerunning{Transformer-based classification of variable-length volumetric scans}


\author{
Marzieh Oghbaie\inst{1,2}
\orcidID{0000-0001-7391-1612}
\and
Teresa Ara\'ujo\inst{1,2}
\orcidID{0000-0001-9687-528X}
\and
Taha Emre\inst{2}
\orcidID{0000-0002-6753-5048}
\and
Ursula Schmidt-Erfurth\inst{1}
\orcidID{0000-0002-7788-7311}
\and
Hrvoje Bogunovi\'c\inst{1,2}
\orcidID{0000-0002-9168-0894}
}

%index{Oghbaie, Marzieh}
%index{Ara\'ujo, Teresa}
%index{Emre, Taha}
%index{Schmidt-Erfurth, Ursula}
%index{Bogunovi\'c, Hrvoje}

%
\authorrunning{M. Oghbaie et al.}
\institute{
Christian Doppler Laboratory for Artificial Intelligence in Retina, Department of Ophthalmology and Optometry, Medical University of Vienna, Austria \and
Lab for Ophthalmic Image Analysis, Departmemnt of Ophthalmology and Optometry, Medical University of Vienna, Austria\\
\email{\{marzieh.oghbaie, hrvoje.bogunovic\}@meduniwien.ac.at}
}
\maketitle              % typeset the header of the contribution
\begin{abstract}
The automatic classification of 3D medical data is memory-intensive. Also, variations in the number of slices between samples is common. Naïve solutions such as subsampling can solve these problems, but at the cost of potentially eliminating relevant diagnosis information. Transformers have shown promising performance for sequential data analysis.  However, their application for long sequences is data, computationally, and memory demanding.
In this paper, we propose an end-to-end Transformer-based framework that allows to classify volumetric data of variable length in an efficient fashion. 
Particularly, by randomizing the input volume-wise resolution(\#slices) during training, we enhance the capacity of the learnable positional embedding assigned to each volume slice. Consequently, the accumulated positional information in each positional embedding can be generalized to the neighbouring slices, even for high-resolution volumes at the test time. 
By doing so, the model will be more robust to variable volume length and amenable to different computational budgets.
We evaluated the proposed approach in retinal OCT volume classification and achieved 21.96$\%$ average improvement in balanced accuracy on a 9-class diagnostic task, compared to state-of-the-art video transformers. 
Our findings show that varying the volume-wise resolution of the input during training results in more informative volume representation as compared to training with fixed number of slices per volume. 

\keywords{Optical coherence tomography  \and 3D volume classification \and Transformers}
\end{abstract}

\section{Introduction}\label{introduction}

% Motivation (generic) =============================================
Volumetric medical scans allow for comprehensive diagnosis, but their manual interpretation is time consuming and error prone \cite{santos2018semivariogram,singh20203d}. Deep learning methods have shown exceptional performance in automating this task \cite{wulczyn2020deep}, often at medical expert levels \cite{de2018clinically}. However, their application in the clinical practice is still limited, partially because they require rigid acquisition settings.
In particular, variable volume length, i.e. number of slices, is common for imaging modalities such as computed tomography, magnetic resonance imaging or optical coherence tomography (OCT). Despite the advantages of having data diversity in terms of quality and size, automated classification of dense scans with variable input size is a challenge. 
Furthermore, the 3D nature of medical volumes results in a memory-intensive training procedure when processing the entire volume.
To account for this constraint and make the input size uniform, volumes are usually subsampled, ignoring and potentially hiding relevant diagnostic information.

% State-of-the-art for variable length input ====================
Among approaches for handling variable input size, Multiple Instance Learning (MIL) is commonly used. There, a model classifies each slice or subgroup of slices individually, and the final prediction is determined by aggregating sub-decisions via maximum or average pooling \cite{rasti2017macular,de2021making,qiu2019self}, or other more sophisticated fusion approaches \cite{simonyan2014two,wang2018video}.
However, they often do not take advantage of the 3D aspect of the data.
The same problem occurs when stacking slice-wise embeddings \cite{sun2020automatic,howard2020improving,chung2017lip}, applying self-attention \cite{das2020b} for feature aggregation, or using principal component analysis (PCA) \cite{fang2017automatic} to reduce the variable number of embeddings to a fixed size.
As an alternative, recurrent neural networks (RNNs) \cite{romo2020end} consider the volume as a sequence of arranged slices or the corresponding embeddings. However, their performance is overshadowed by arduous training and lack of parallelization.

Vision Transformers (ViTs) \cite{dosovitskiy2020image}, on the other hand, allow parallel computation and effective analysis of longer sequences by benefiting from multi-head self-attention (MSA) and positional encoding. These components allow to model both local and global dependencies, playing a pivotal role for 3D medical tasks where the order of slices is important \cite{windsor2022context,peiris2022robust,playout2022focused}.
Moreover, ViTs are more flexible regarding input size. Ignoring slice positional information (bag-of-slices) or using sinusoidal positional encoding enables them to process sequences of arbitrary length with respect to computational resources. However, ViTs with learnable positional embeddings (PEs) have shown better performance \cite{dosovitskiy2020image}. In this case, the only restriction in processing variable length sequences is the number of PEs. Although interpolating the PE sequence helps overcome this restriction, the resultant sequence will not model the exact positional information of the corresponding slices in the input sequence, affecting ViTs performance \cite{beyer2022flexivit}.
Notably, Flexible ViT~{\cite{beyer2022flexivit}} (FlexiViT) handles patch sequences of variable sizes by randomizing the patch size during training and, accordingly, resizing the embedding weights and parameters corresponding to PEs. 

% Contributions =================================================
Despite the merits of the aforementioned approaches, three fundamental challenges still remain.
First, the model should be able to process inputs with variable  volume resolutions, where throughout the paper we refer to the resolution in the dimension across slices (number of slices), and simultaneously capture the size-independent characteristics of the volume and similarities among the constituent slices. The second challenge is the scalability and the ability of the model to adapt to unseen volume-wise resolutions at inference time. Lastly, the training of deep learning models with high resolution volumes is both computationally expensive and memory-consuming. 

In this paper, we propose a late fusion Transformer-based end-to-end framework for 3D volume classification whose local-similarity-aware PEs not only improve the model performance, but also make it more robust to interpolation of PEs sequence. 
We first embed each slice by a spatial feature extractor and then aggregate the corresponding sequence of slice-wise embeddings with a Feature Aggregator Transformer (FAT) module to capture 3D intrinsic characteristics of the volume and produce a volume-level representation. To enable the model to process volumes with variable resolutions, we propose a novel training strategy, Variable Length FAT (VLFAT), that enables FAT module to process volumes with different resolutions both at training and test times. VLFAT can be trained with a proportionally few \#slices, an efficient trait in case of training time/memory constraints. Consequently, even with drastic slice subsampling during training, the model will be robust against extreme PEs interpolation for high-resolution volumes at the test time.
The proposed approach is model-agnostic and can be deployed with Transformer-based backbones.
VLFAT beats the state-of-the-art performance in retinal OCT volume classification on a private dataset with nine disease classes, and achieves competitive performance on a two-class public dataset.
\section{Methods}
Our end-to-end Transformer-based volume classification framework (Fig.~\ref{fig:proposed_method}) has three main components: 
1) Slice feature extractor (SFE) to extract spatial biomarkers and create a representation of the corresponding slice; 2) Volume feature aggregator (VFA) to combine the slice-level representations into a volume-level representation, and
3) Volume classification.
Trained with the proposed strategy, our approach is capable of processing and classifying volumes with varying volume-wise resolutions.
Let's consider a full volume \(v\in \mathbf{R}^{(N \times W \times H)}\), where (N, H, W) are the \#slices, its width and height respectively. The input to the network is a subsampled volume by randomly selecting \(n\) slices.

\subsubsection{Slice feature extractor (SFE)} 
To obtain the slice representations, we use ViT as our SFE due to its recent success in medical interpretation tasks \cite{he2022transformers}. ViT mines crucial details from each slice and, using MSA and PE, accumulates the collected information in a learnable classification token, constituting the slice-wise embedding. 
For each slice token, we then add a learnable 1D PE \cite{dosovitskiy2020image} to retain the position of each slice in the volume.
\subsubsection{Volume feature aggregator (VFA)} 
\begin{figure*}[tb]
    \centering    
    \includegraphics[height=4.5cm]{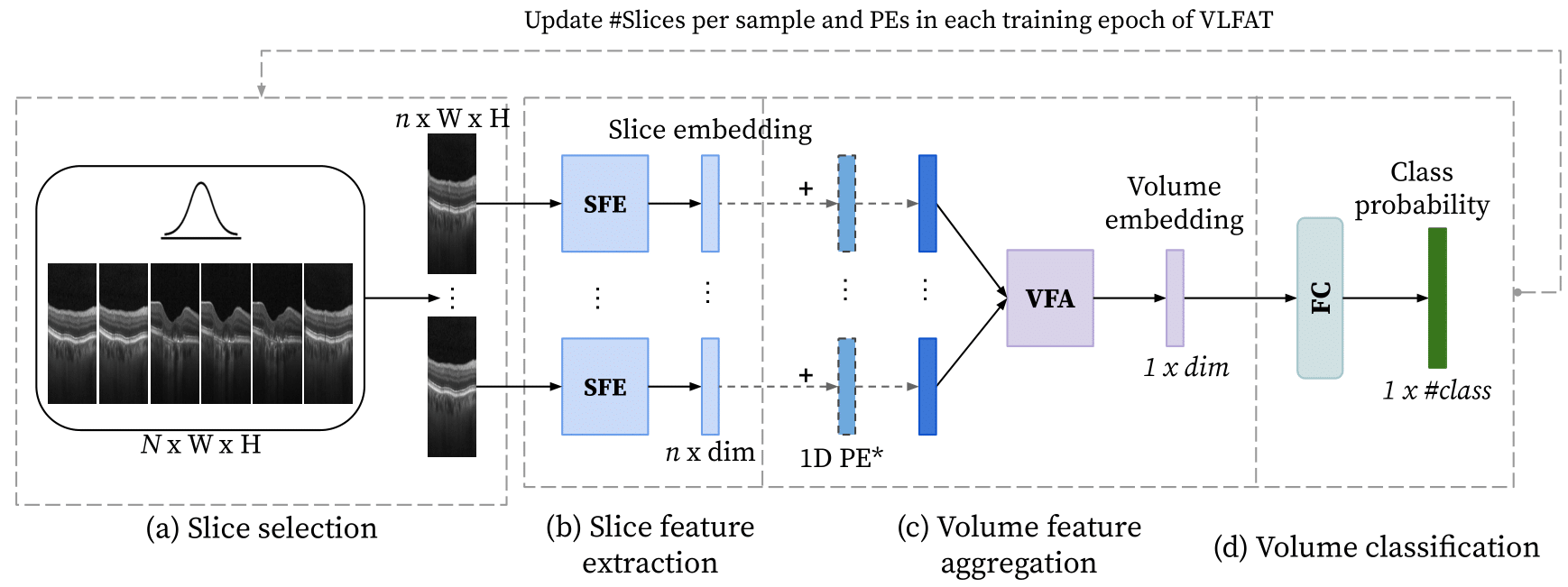}
    \caption{The overview of the proposed Transformer-based approach for 3D volume classification. The shared SFE processes the input slices, and in line with VLFAT, both \#slices and the PEs sequence are updated at each epoch. $^*$1D PE is added to each slice embedding for FAT and VLFAT.}
    \label{fig:proposed_method}
\end{figure*}

The output of the previous step is a sequence of slice-wise embeddings, to which we append a learnable volume-level classification token~{\cite{devlin2018bert}}.
The resulting sequence of embedding vectors is then processed by the FAT module to produce a volume-level embedding. 
In particular, we propose 
\textit{VLFAT}, a FAT with enhanced learnable PEs, inspired on FlexiViT \cite{beyer2022flexivit}, where we modify \#slices per input volume instead of patch sizes and correspondingly apply PEs interpolation. This allows handling arbitrary volume resolutions, which generally would not be possible except for an ensemble of models of different scales.
Specifically, at initialization we set a fixed value, \(n\), for \#slices, resulting in PEs sequence with size \((n+1, dim)\), where an extra PE is assigned to the classification token and \(dim\) is the dimension of the slice representation. In each training step, we then randomly sample a new value for \(n\) from a predefined set and, accordingly, linearly interpolate the PEs sequence (Fig.~\ref{fig:proposed_method}), using the known adjacent PEs~\cite{blu2004linear}. This allows to preserve the similarity between neighboring slices in the volume, sharing biomarkers in terms of locality, and propagating the corresponding positional information. The new PEs are then normalized according to a truncated normal distribution.

\subsubsection{Volume classification} Finally, the volume-level classification token is fed to a Fully Connected (FC) layer, which produces individual class scores. As a loss function, we employ the weighted cross-entropy.

% ===============================================================
% ===============================================================

\section{Experiments}\label{experiments}

We tested our model for volume classification of macula-centered retinal OCT scans, where large variation in volume resolution (\#B-scans) between samples is very common. For multiclass classification performance metrics, we relied on Balanced Accuracy (BAcc) and one-vs-all Area Under the Receiver Operating Curve (AUROC). The source code is available at: github.com/marziehoghbaie/VLFAT.

\subsubsection{Datasets} We utilized three large retinal OCT volumetric datasets: \textit{Duke} for pre-training all models, and \textit{9C} and \textit{OLIVES} for fine-tuning and testing purposes.

\begin{itemize}
    \item \textit{Duke:} Public dataset \cite{sun2020automatic} with 269 intermediate age-related macular degeneration (iAMD) and 115 normal patients, acquired with a Bioptigen OCT device with a resolution of 100 B-scans per volume. Volumes were split patient-wise into 80\% for training (384 samples) and 20\% for validation (77 samples). 
    \item \textit{9C:} \label{section_datasets} Private dataset with 4766 volumes (4711 patients) containing 9 disease classes: iAMD, three types of choroidal neovascularization (CNV1-3), geographic atrophy (GA), retinal vein occlusion (RVO), diabetic macular edema (DME), Stargardt disease, and healthy. Volumes were split patient-wise, for each class, into 70\% training (3302 samples), 15\% validation (742 samples), and 15\% test (722 samples).
The OCT volumes were acquired by four devices (Heidelberg Engineering, Zeiss, Topcon, Nidek), exhibiting large variation in \#slices. Minimum, maximum, and average \#slices per volume were 25, 261, and 81, respectively. 
    \item \textit{OLIVES:} Public dataset \cite{prabhushankar2022olives} with 3135 volumes (96 patients) labeled as diabetic retinopathy (DR) or DME. OCTs were acquired with Heidelberg Engineering device, and have resolutions of either 49 or 97, and were split patient-wise for each class into 80\% training (1808 samples), 10\% validation (222 samples), and 10\% test (189 samples). 
\end{itemize}

\begin{figure*}[tb]
    \centering    
    \includegraphics[height=4.5cm]{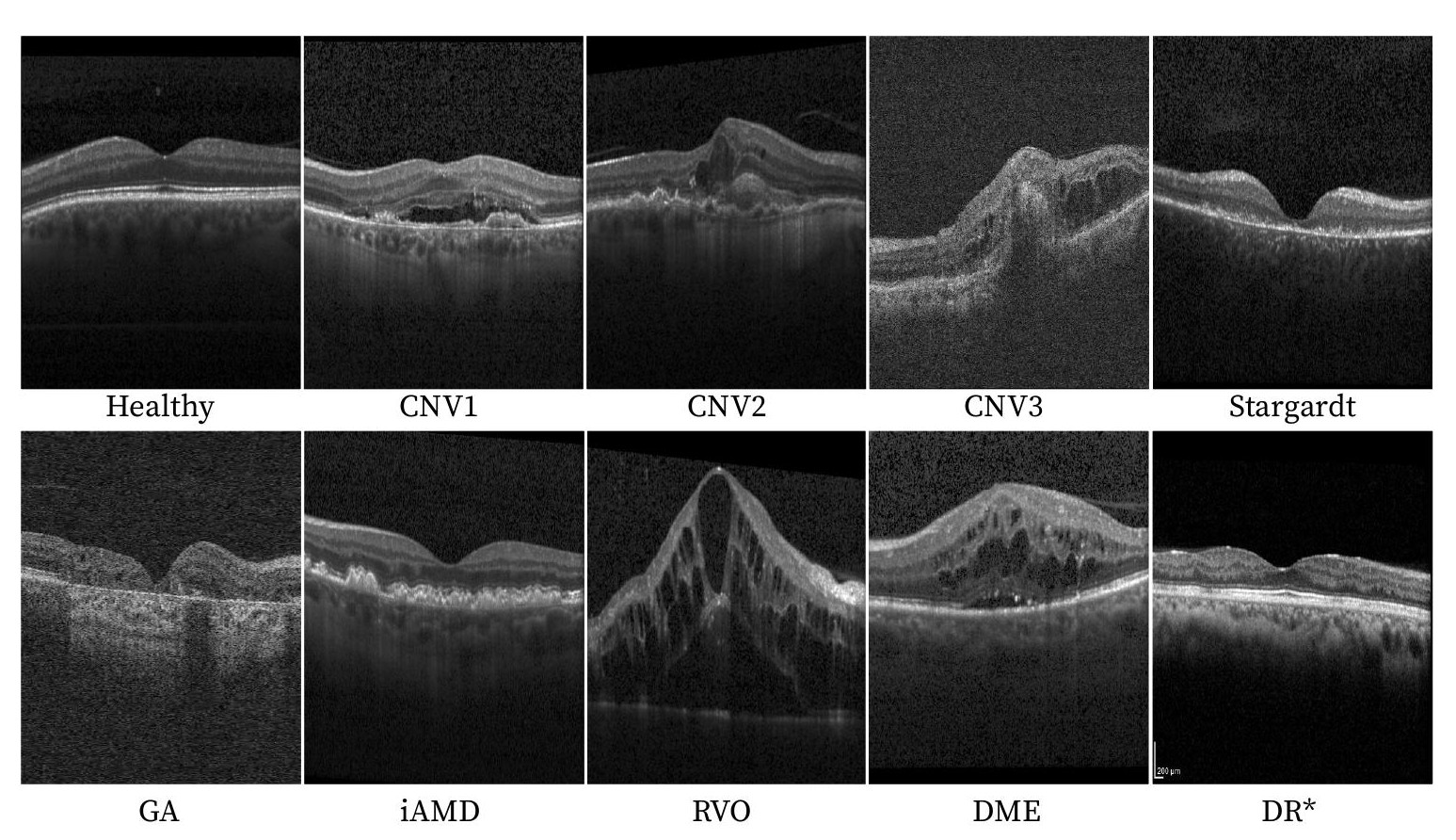}
    \caption {Samples of central B-scan from all disease classes. $^*$DR is only in OLIVES.}
    \label{fig:samples}
\end{figure*}

\subsubsection{Comparison to state-of-the-art methods}
We compared the performance of the proposed method with two state-of-the-art video ViTs (ViViT) \cite{arnab2021vivit}, originally designed for natural video classification: 
1) factorized encoder (FE) ViViT, that models the spatial and temporal dimensions separately; 2) factorised self-attention (FSA) ViViT, that simultaneously computes spatial and temporal interactions.
We selected FE and FSA ViViTs as baselines to understand the importance of separate feature extractors and late fusion in our approach. FE ViViT, similar to ours, utilizes late fusion, while FSA ViViT is a slow-fusion model and processes spatiotemporal patches as tokens. 

\subsubsection{Ablation studies} 
To investigate the contribution of SFE module, we deployed ViT and ResNet18~\cite{windsor2022context}, a standard 2D convolutional neural network (CNN) in medical image analysis, with pooling methods as VFA where the quality of slice-wise features is more inﬂuential. For VFA, we explored average pooling (AP), max pooling (MP), and 1D convolution (1DConv). As MIL-based baselines, pooling methods can be viable alternatives to VLFAT for processing variable volume resolutions. In addition to learnable PE, we deployed sinusoidal PE (sinPE) and bag-of-slices (noPE) for FAT to examine the effect of positional information.

\subsubsection{Robustness analysis} 
We investigate the robustness of VFLAT and FAT to PEs sequence interpolation at inference time by changing the volume resolution. To process inputs with volume resolutions different from FAT's and VLFAT's input size, we linearly interpolate the sequence of PEs at the test time. For 9C dataset, we only assess samples with minimum \#slices of 128 to better examine the PE's scalability to higher resolutions.
\paragraph{Implementation details}
The volume input size was $25\times224\times224$ for all experiments except for FSA ViViT where \#slices was set to 24 based on the corresponding tublet size of \(2 \times 16 \times 16\). During VLFAT training, the \#slices varied between $\{5, 10, 15, 20, 25\}$, specified according to memory constraints.   
We randomly selected slices using a normal distribution with its mean at the central slice position, thus promoting the inclusion of the region near the fovea, essential for the diagnosis of macular diseases. Our ViT configuration is based on ViT-Base \cite{dosovitskiy2020image} with patch size \(16 \times 16\), and 12 Transformer blocks and heads. For FAT and VLFAT, we set the number of Transformer blocks to 12 and heads to 3. The slice-wise and volume-wise embedding dimension were set to 768. The configuration of ViViT baselines was set according to the original papers \cite{arnab2021vivit}. Training was performed using AdamW \cite{loshchilov2017decoupled} optimizer with learning rate of $6\times10^{-6}$ with cosine annealing. All models were trained for 600 epochs with a batch size of 8.
Data augmentation included random brightness enhancing, motion blur, salt/pepper noise, rotation, and random erasing~\cite{zhong2020random}.
%\par
The best model was selected based on the highest BAcc on the validation set. 
All experiments were performed using Pytorch 1.13.0+cu117 and timm library~\cite{rw2019timm} on a server with 1TB RAM, and NVIDIA RTX A6000 (48GB VRAM).

\section{Results and Discussion}
% State of the art comparison  ------------
In particular, on large 9C dataset our VLFAT achieved 21.4\% and 22.51\% BAcc improvement compared to FE ViViT and FSA ViViT, respectively.
% Ablation on VFA ------------
Incorporating our training strategy, VLFAT, improved FAT's performance by 16.12\% on 9C, and 8.79\% on OLIVES, which verifies the ability of VLFAT in learning more location-aware PEs, something that is also reflected in the increase of AUROCs (0.96\textrightarrow 0.98 on 9C dataset and 0.95\textrightarrow 0.97 on OLIVES).
% Ablation on PE ------------
Per-class AUROCs are shown in Table~\ref{tab:AUC_9class}. The results show that for most of the classes, our VLFAT has better diagnostic ability and collects more disease-specific clues from the volume. 

\begin{figure*}[tb]
    \centering
    \includegraphics[height=4cm]{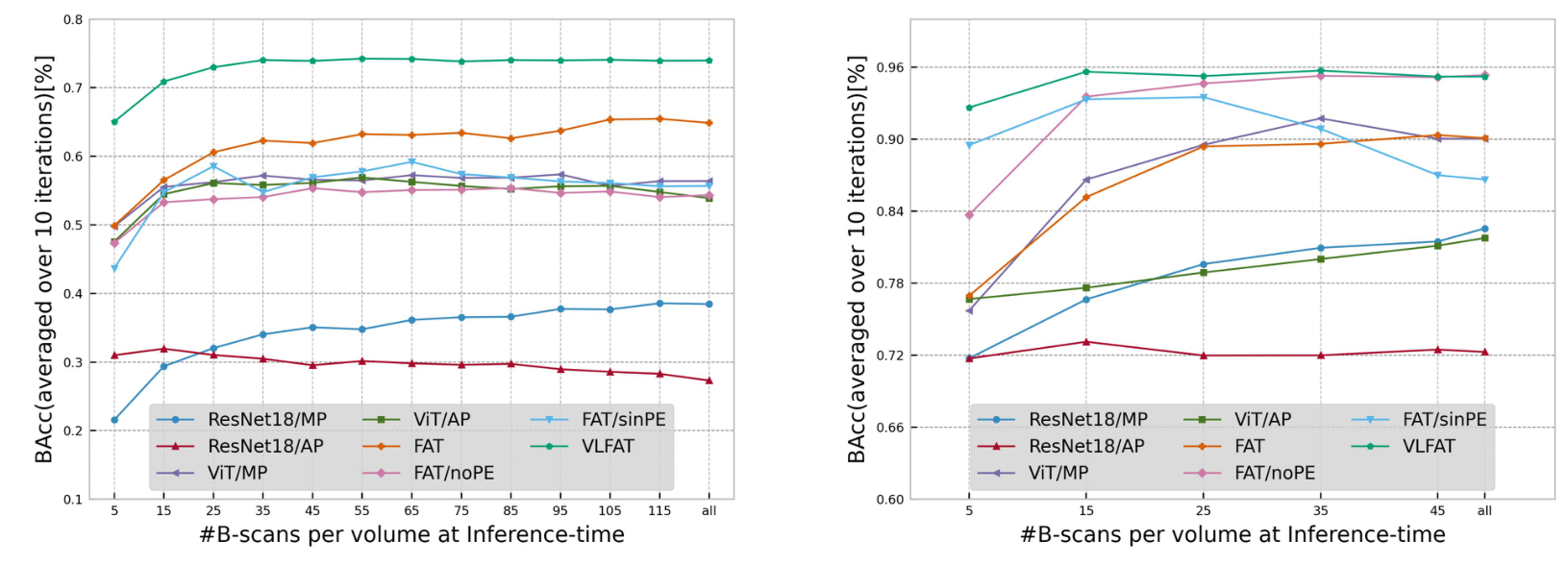}
    \caption{ Robustness analysis of VLFAT and vanilla FAT against PEs interpolation at the test time: (a) 9C dataset; (b) OLIVES} 
    \label{fig:robustness}
\end{figure*}

% Ablation studies (generic) ----------
The ablation study (Table~\ref{tab:res_9class}) showed that each introduced component in the proposed model contributed to the performance improvement. In particular, ViT was shown as a better slice feature extractor compared to ResNet18, particularly on 9C dataset where the differences between disease-related biomarkers are more subtle. 
Additionally, the poor performance of the pooling methods as compared to FAT and 1DConv, emphasizes the importance of contextual volumetric information,
the necessity of a learnable VFA, and the superiority of Transformers over 1DConv.
% ====== Results of OLIVES and ablation over PE======
Although, on OLIVES, less complicated VFAs (pooling/1DConv) and  FAT (noPE) also achieved comparable results, which can be attributed primarily to DR vs. DME~\cite{prabhushankar2022olives} being an easier classification task compared to the diverse disease severity in the 9C dataset.
In addition, the competitive advantage of VLFAT in handling different resolutions was not fully exploited in OLIVES since the large majority of cases had the same \#slices.
On 9C, however, the comparison of positional encoding strategies demonstrated that although ignoring PEs and sinusoidal approach provide deterministic predictions, the importance of learnable PEs in modeling the anatomical order of slices in the volume is crucial.
% Robustness experiment ------------
The robustness analysis is shown in Fig.~\ref{fig:robustness}. VLFAT was observed to have more scalable and robust PEs when the volume-wise resolutions at the test time deviated from those used during training. This finding highlights the VLFAT's potential for resource-efficient training and inference.

\begin{table}
    \centering
    \caption{Classification performance in terms of balanced accuracy (BAcc) and mean one-vs-all AUROC.
    FE: Factorised Encoder, FSA: Factorised Self-Attention, SFE: Slice Feature Extractor, VFA: Volume Feature Aggregator.
    }
    \label{tab:res_9class}
    \begin{tabular}[width\textwidth]{l|c|c|c|c|c} 
    \hline
    Method (SFE/VFA) & \multicolumn{2}{|c|}{9C} & \multicolumn{2}{|c|}{OLIVES}& $\#$slices$^*$\\
    \hline
    & BAcc & AUROC & BAcc & AUROC & \\
    \hline
    FE ViViT (baseline)\cite{arnab2021vivit} & 0.64 & 0.96  & 0.93 & 0.98 & 25\\
    FSA ViViT (baseline)\cite{arnab2021vivit} & 0.63 & 0.95 & 0.92 & 0.98 & 24\\
    ViT/1DConv & 0.61 & 0.94 & 0.95 & 0.97 & 25\\
    ResNet18/AP & 0.34 & 0.73 & 0.72 & 0.83 & all\\
    ResNet18/MP & 0.41 & 0.87 & 0.83 & 0.93 & all\\
    ViT/AP & 0.59 & 0.95 & 0.82 & 0.97 & all\\
    ViT/MP& 0.61 & 0.96 & 0.90 & 0.98 & all\\
    ViT/FAT (noPE) & 0.58 & 0.93 & \textbf{0.95}& \textbf{0.99}  & all\\
    ViT/FAT (sinPE) & 0.62 & 0.93 & 0.87 & 0.96 & all\\
    ViT/FAT$^+$ & 0.67 & 0.96 & 0.88 & 0.95  & 25\\
    ViT/VLFAT (ours) & \textbf{0.78} & \textbf{0.98} & 0.95 & 0.97 & all\\
    \hline
    \multicolumn{6}{l}{\scriptsize Legend: $^*$\#slices at the test time;}\\
    \multicolumn{6}{l}{\scriptsize $^+$the input length is fixed in both training and test time}
\end{tabular}
\end{table}

\begin{table}
    \centering
    \caption{Per-class classification performance (one-vs-all AUROC) on 9C dataset.}
    \label{tab:AUC_9class}
    \begin{tabular}[width\textwidth]{l|c|c|c|c|c|c|c|c|c}
    \hline
    Method(SFE/VFA) & CNV1& CNV2 & CNV3 & DME& GA& Healthy & iAMD & RVO & Stargardt\\
    \hline
    FE ViViT\cite{arnab2021vivit} &0.93 & 0.91& 0.95& 0.94& 0.99& 0.95& 0.92& 0.95&0.99\\
    FSA ViViT\cite{arnab2021vivit} & 0.94& 0.91& 0.92& 0.92& \textbf{1.0}&0.94 &0.93 &0.94 &0.99\\
    ViT/1DConv &0.88 & 0.92& 0.94& 0.91& 0.98& 0.92& 0.92& 0.92&\textbf{1.0}\\
    ResNet18/AP & 0.68 & 0.63 & 0.58 & 0.75 & 0.81 & 0.75 & 0.75 & 0.76 & 0.97\\
    ResNet18/MP & 0.78 & 0.77& 0.79& 0.87& 0.91& 0.84& 0.84& 0.84&0.91\\
    ViT/AP  & 0.9& 0.81& 0.92& 0.93& 0.98& 0.94& 0.93& 0.94&0.99\\
    ViT/MP  &0.95 & 0.85& 0.95& 0.94& 0.98& 0.94& 0.93& 0.96&0.99\\
    ViT/FAT (noPE) &0.9 &0.87 & 0.89& 0.89& 0.98& 0.91& 0.9& 0.94&0.99\\
    ViT/FAT (sinPE) &0.94 &\textbf{0.93} & 0.93& 0.88& 0.97& 0.91& 0.89& 0.93&0.99\\
    ViT/FAT &0.93 & 0.82& 0.97&0.94 &0.99 &0.95& 0.94& 0.95&0.99\\
    ViT/VLFAT (ours) &\textbf{0.98} & 0.92& \textbf{0.98}& \textbf{0.98}& \textbf{1.0}& \textbf{0.99}& \textbf{0.98}&\textbf{0.98} &\textbf{1.0}\\
    \hline
\end{tabular}
\end{table}

\section{Conclusions}
In this paper, we propose an end-to-end framework for 3D volume classification of variable-length scans, benefiting from ViT to process volume slices and FAT to capture 3D information. Furthermore, we enhance the capacity of PE in FAT to capture sequential dependencies along volumes with variable resolutions. Our proposed approach, VLFAT, is more scalable and robust than vanilla FAT at classifying OCT volumes of different resolutions.
On a large-scale retinal OCT datasets, our results indicate that this effective method performs in the majority of cases better than other common methods for volume classification. 

Besides its applicability for volumetric medical data analysis, our VFLAT has potential to be applied on other medical tasks including video analysis (e.g. ultrasound videos) and high-resolution imaging, as is the case in histopathology.
Future work would include adapting VLFAT to ViViT models to make them less computationally expensive. 
Furthermore, PEs in VLFAT could be leveraged for improving the visual interpretation of decision models by collecting positional information about the adjacent slices sharing anatomical similarities.

\subsubsection{Acknowledgements.}%
This work was supported in part by the Christian Doppler Research Association, Austrian Federal Ministry for Digital and Economic Affairs, the National Foundation for Research, Technology and Development, and Heidelberg Engineering.

\bibliographystyle{splncs04}
\bibliography{References}

\end{document}